\documentclass[journal]{IEEEtran}
\ifCLASSINFOpdf
\else
\fi
\usepackage{microtype}
\usepackage{graphicx}
\usepackage{booktabs} % for professional tables
\usepackage[utf8]{inputenc} % allow utf-8 input
\usepackage[T1]{fontenc}    % use 8-bit T1 fonts
\usepackage{url}            % simple URL typesetting
\usepackage{booktabs}       % professional-quality tables
\usepackage{amsfonts}       % blackboard math symbols
\usepackage{nicefrac}       % compact symbols for 1/2, etc.
\usepackage{microtype}      % microtypography

\usepackage{amsmath}
\usepackage{amsthm}
\usepackage{amssymb}
\usepackage{xcolor}
\usepackage{bm}
\usepackage{graphicx}
\usepackage{bbm}

\def \mn{(\mu,\nu)}

\def \PP{\mathcal{P}}
\def \ff{\mathbb{F}}

\def \smallceilN{\frac{N}{2}}

\DeclareMathOperator*{\argmax}{argmax}

\DeclareMathOperator*{\Real}{Re}

\DeclareMathOperator*{\diag}{diag}

\definecolor{zambocolor}{RGB}{0, 153, 51}

\def \KK{\mathbb{K}}
\def \const{C}
\newtheorem{example}{Example}
\newtheorem{theorem}{Theorem}
\newtheorem{proposition}{Proposition}
\newtheorem{definition}{Definition}
\newtheorem{corollary}{Corollary}
\newtheorem{lemma}[theorem]{Lemma}
\newtheorem{remark}{Remark}

\newtheorem{conjecture}{Conjecture}

\begin{document}
\title{The Fourier Discrepancy Function}

% \author{Gennaro~Auricchio,~\IEEEmembership{Member,~IEEE,}
%         Andrea~Codegoni,~\IEEEmembership{Fellow,~OSA,}Stefano~Gualandi,~\IEEEmembership{Fellow,~OSA,} 
%         and~Lorenzo~Zambon,~\IEEEmembership{Life~Fellow,~IEEE}% <-this % stops a space
\author{Gennaro~Auricchio,
        Andrea~Codegoni, Stefano~Gualandi 
        and~Lorenzo~Zambon% <-this % stops a space
\thanks{G. Auricchio, A. Codegoni, S. Gualandi, and L. Zambon are with the Department
of Mathematics, University of Pavia, Pavia,
PV, 27100 Italy e-mail: gennaro.auricchio@unipv.it, andrea.codegoni01@universitadipavia.it, stefano.gualandi@unipv.it, lorenzogianmar.zambon01@universitadipavia.it
%(see http://www.michaelshell.org/contact.html).
}% <-this % stops a space
% \thanks{J. Doe and J. Doe are with Anonymous University.}% <-this % stops a space
% \thanks{Manuscript received April 19, 2005; revised August 26, 2015.}
}

\maketitle

% We also show that this discrepancy takes into account the geometry of the underlying space. We prove that this discrepancy enjoys many desirable features, such as convexity. We also show that this discrepancy takes into account the geometry of the underlying space. Moreover, we relate the minimisation of the Fourier Discrepancy to the maximum likelihood estimator in classification models with a Gaussian noise in the space of frequencies.

% \genna{GRAMMARLY SUGGERISCE:}

% \genna{This paper proposes the Fourier Discrepancy Function, a new discrepancy to compare discrete probability measures. 
% We show that this discrepancy takes into account the geometry of the underlying space. 
% We prove that the Fourier Discrepancy is convex, twice differentiable, and that its gradient has an explicit formula.
% We also provide a compelling statistical interpretation.
% Finally, we study the lower and upper tight bounds for the Fourier Discrepancy in terms of the Total Variation distance.}
%We relate this discrepancy to the most commonly used by providing theoretical upper bounds.
%We prove that it is dominated by the most commonly used discrepancies.
%We show that this discrepancy is twice differentiable, strictly convex and that its gradient has an explicit formula.
%Finally, we relate the minimization of the Fourier Discrepancy to the maximum likelihood estimator in classification models with a Gaussian noise in the space of frequencies.

\begin{abstract}

In this paper, we propose the Fourier Discrepancy Function, a new discrepancy to compare discrete probability measures. 
We show that this discrepancy takes into account the geometry of the underlying space. 
We prove that the Fourier Discrepancy is convex, twice differentiable, and that its gradient has an explicit formula.
We also provide a compelling statistical interpretation.
Finally, we study the lower and upper tight bounds for the Fourier Discrepancy in terms of the Total Variation distance.

\end{abstract}


% \fi
\IEEEpeerreviewmaketitle

\section{Introduction}
% for IEEE journal papers produced under \LaTeX\ using
% IEEEtran.cls version 1.8b and later.
% % You must have at least 2 lines in the paragraph with the drop letter
% % (should never be an issue)
% I wish you the best of success.

 

% \subsection{Subsection Heading Here}
% Subsection text here.

% % needed in second column of first page if using \IEEEpubid
% %\IEEEpubidadjcol

% \subsubsection{Subsubsection Heading Here}
% Subsubsection text here.
\IEEEPARstart{C}{omparing} 
%able to compare
probability measures is 
a crucial task in several applied fields, such as computer vision \cite{haker2004optimal,Bonneel2019,auricchio2018computing,bassetti2018computation,auricchio2019computing,cuturi2014fast,papadakis2015optimal}, supervised learning \cite{janocha2017loss,bengio2017deep,bishop2006pattern,schmidhuber2015deep,frogner2015learning} and generative models \cite{arjovsky2017wasserstein,ansari2020characteristic,li2017mmd}.
%image retrieval \cite{rubner2000earth}, histogram comparison in higher dimensions \cite{auricchio2018computing,bassetti2018computation}, image registration \cite{haker2004optimal,Bonneel2019}, computation of barycenters among images \cite{auricchio2019computing,cuturi2014fast}, and supervised learning \cite{janocha2017loss}.
However, using different metrics for a given task can lead to different results \cite{arjovsky2017wasserstein}. 
For this reason, it is of crucial importance to have a wide range of mathematical tools and understand their features. 
For instance, in \cite{lin1991divergence}, a class of divergence measures based on the Shannon entropy has been introduced and studied.
A relevant topic in information theory has then become to give a comparison between different discrepancy functions, especially in terms of tight bounds. 
The problem of finding tight bounds has been introduced in \cite{gilardoni2006minimum}; since then, many works have improved the constants of several known inequalities \cite{topsoe2000some,guntuboyina2013sharp,gilardoni2010pinsker}. These bounds have also been proved to be useful for source coding \cite{csiszar1967two,csiszar1967information,sason2014tight}.

In this paper, we introduce the Fourier Discrepancy Function, a distance between discrete probability measures inspired by the $1,2-$Periodic Fourier-Based Metric \cite{auricchio2020equivalence}. Metrics based on the Fourier Transform have been introduced in \cite{gabetta1995metrics} and used in several fields, such as kinetic theory \cite{carrillo2007contractive,baringhaus1997class}, statistics \cite{heathcote1977integrated}, and, more recently, generative models \cite{ansari2020characteristic}. 
The Fourier Discrepancy inherits the ability to capture the geometry of the underlying space, which is an appealing property in several applications \cite{rubner2000earth,ruzon2001edge,peyre2019computational}.  Moreover, it is easy to compute using the Fast Fourier Transform \cite{cooley1969fast}.

%Metrics based on the Fourier Transform have been introduced in \cite{gabetta1995metrics} and used in several fields, such as kinetic theory \cite{carrillo2007contractive,baringhaus1997class}, statistics \cite{heathcote1977integrated}, and, more recently, generative models \cite{ansari2020characteristic}. 

%The paper is organized as follows. 
In Section \ref{CULF}, we recall the most commonly used discrepancy functions for discrete probability measures and the $1,2-$Periodic Fourier Based Metric \cite{auricchio2020equivalence}. 
Then, in Section \ref{thisd}, we introduce the Fourier Discrepancy.
%and show its analytical properties and a nice statistical interpretation.
We prove that the squared Fourier Discrepancy is twice differentiable. Unlike the Wasserstein distance \cite{peyre2015entropic,levy2018notions}, its gradient has an explicit formula. 
%We also prove that the Fourier Discrepancy is convex.
%Then, we show that this discrepancy has an interesting statistical interpretation. Indeed, if we deal with a Gaussian noise in the space of frequencies, the minimisation of the Fourier Discrepancy is equivalent to the maximisation of the likelihood of the observations.
Moreover, we prove that the Fourier Discrepancy is convex, and we provide an interesting statistical interpretation.
Finally, in Section \ref{section tight bounds}, we study the lower and upper tight bounds for the Fourier Discrepancy in terms of the Total Variation distance. 
We close our paper with an open conjecture on the value of the upper tight bound.

\section{Discrepancy Functions for Probability Measures}
\label{CULF}
% \code{Per me discrepancy è sbagliato, meglio divergence}
%In this section, we set our notation and recall the main mathematical tools that will be used throughout the paper.
\subsection{Commonly used discrepancy functions}

In this subsection, we recall the main distances and divergences used to compare discrete probability measures in computational applied mathematics.
Here $(X,d)$ denotes a discrete finite metric space, and $\PP(X)$ denotes the set of all the probability measures over $X$. For a complete discussion, we refer to \cite{ccinlar2011probability,villani2008optimal}.

% \begin{definition}
% Given a target measure $\nu\in\PP(X)$ and a function $\Psi:\PP(X)\times \PP(X)\to \erre$, we define the discrepancy of $\mu$ (with respect to $\nu$) as
% \begin{equation}
%     \label{eq:generic_loss}
%     L_\nu^{(\Psi)}(\mu):=\Psi(\mu,\nu).
% \end{equation}
% \end{definition}
% The role of the discrepancy function is to quantify the differences between the measure $\mu$ and the target $\nu$, therefore, the function $\Psi$ is usually a distance.

\begin{itemize}
    \item The Total Variation distance ($TV$) 
    \cite{ccinlar2011probability}
    is defined as
    \[
    TV\mn:=\dfrac{1}{2}\sum_{x\in X}|\mu_x-\nu_x|.
    \]
    
    \item The Kullback-Leibler divergence ($KL$) \cite{kullback1951information} is defined as
    %, with the convention $0\cdot\log(0)=0$, as
    \begin{equation}
        KL(\nu||\mu):=
    \sum_{x\in X}\log\Big(\dfrac{\nu_x}{\mu_x}\Big)\nu_x,
    \label{eq:kl}
    \end{equation}

   if $\nu_x = 0$ for every $x$ such that $\mu_x = 0$, and otherwise $KL(\nu||\mu):= +\infty$.
   We follow the convention $0\cdot\log(0)=0$.
  
% ALTERNATIVA:   
%   \begin{equation}
%        KL(\nu||\mu):= \begin{cases}
%    \sum_{x\in X}\log\Big(\dfrac{\nu_x}{\mu_x}\Big)\nu_x \qquad if \nu_x = 0 \forall x s.t. \mu_x = 0\\
%    +\infty \qquad otherwise,
%    \end{cases}
%    \label{eq:kl}
%    \end{equation}
   
%   In classification tasks the  label $\nu$ is fixed. Therefore we can simplify equation \eqref{eq:kl} by removing the constant term $\sum_{x\in X}\nu_x\log(\nu_x)$. As a result, we obtain the Cross Entropy ($XE$), which is defined as
%     \begin{equation}
%         XE(\mu,\nu):=-\sum_{x\in X} \nu_x\log(\mu_x).
%         \label{eq:xe}
%     \end{equation}
%     Hence, in classification tasks minimising the Cross Entropy is equivalent to minimising the Kullback-Leibler divergence \cite{bengio2017deep,murphy2012machine}.

    \item The Wasserstein distance ($W_1$) \cite{villani2008optimal,santambrogio2015optimal} is defined as
\begin{equation}
    W_1\mn:=\min_{\pi\in\Pi\mn}\bigg\{\sum_{(x,y)\in X\times X}|x-y| \; \pi_{x,y}\bigg\},
\end{equation}
where

\begin{align*}
   \Pi\mn:=\bigg\{\pi\in&\PP(X\times X) \; \text{s.t.}\\
&\sum_{y\in X}\pi_{x,y}=\mu_x,\; \sum_{x\in X}\pi_{x,y}=\nu_y\bigg\}. 
\end{align*}
% \small
% \begin{equation}
%     \displaystyle{\Pi\mn:=\bigg\{\pi\in\PP(X\times X) \; \text{s.t.}\;\sum_{y\in X}\pi_{x,y}=\mu_x,\; \sum_{x\in X}\pi_{x,y}=\nu_y\bigg\}}.
% \end{equation}
% \normalsize
Intuitively, $\pi_{x,y}$ denotes the amount of mass that is moved from the point $x$ to the point $y$ to reshape the configuration $\mu$ into the configuration $\nu$. The cost of moving a unit of mass from $x$ to $y$ is given by $|x-y|$. The Wasserstein distance is then the minimum cost for performing the total reshape.
% When $X$ is a subset of an Euclidean space and $d_{x,y}=|x-y|$, the distance is called $W_1$.
\end{itemize}

\noindent Although 
% all 
these three functions are commonly used to compare measures, 
%their different behaviours reflect dif 
%the way they compare them is different.
%they show a different behaviour when applied to 
their features, and thus their behaviours when applied to a given task, are different.
% A feature of particular interest is the behaviour concerning convergent sequences. 
Studying these features is crucial to choosing the right discrepancy to be used for the given task.
For example, 
the Total Variation is robust against random noise when used as a loss function for classification tasks \cite{ghosh2017robust},
the Kullback-Leibler divergence is closely related to likelihood maximisation \cite{bengio2017deep,bishop2006pattern},
%the convergence induced by the Wasserstein distance $W_1$ turned out to be the one able to mimic the human eye perception, 
while the Wasserstein distance performs well at capturing the geometry of the underlying space \cite{peyre2019computational,villani2008optimal,santambrogio2015optimal}.

\subsection{The $1,2-$Periodic Fourier-based Metric}
\label{sec:Theory}
In this subsection, we review the main notions about the Fourier Transform of discrete measures (DFT) and about Fourier Based Metrics \cite{carrillo2007contractive,auricchio2020equivalence,gabetta1995metrics}. For a complete discussion on the DFT, we refer to \cite{britanak2001transform}.

In what follows, we fix $X=I_N$, where $I_N \subset [0,1]$ is defined as 
\[
I_N:=\bigg\{0,\dfrac{1}{N},\dots,\dfrac{N-1}{N}\bigg\},
\]
for any given $N \in \mathbb{N}$. A discrete measure $\mu$ on $I_N$ is then defined as 
\begin{equation}
    \mu:=\sum_{j=0}^{N-1}\mu_{j}\delta_{\frac{j}{N}},
\end{equation}
where the values $\mu_j$ are non-negative real numbers such that $\sum_{j=0}^{N-1} \mu_j = 1$. Since any discrete measure supported on $I_N$ is fully characterised by the $N-$uple of positive values $(\mu_0,\dots,\mu_{N-1})$, we refer to discrete measures and vectors interchangeably.

\begin{definition}%[Discrete Fourier Trasform]
\label{def:DFT}
The Discrete Fourier Transform (DFT) of $\mu$ is the $N-$dimensional vector $\hat{\mu}:=(\hat{\mu}_0,\dots,\hat{\mu}_{N-1})$ defined as
\begin{equation}
    \hat{\mu}_k:=\sum_{j=0}^{N-1}\mu_{j} e^{-2\pi i\frac{j}{N}k},\quad\quad\quad k \in \{0,\dots,N-1\}.
\end{equation}
\end{definition}

\begin{remark}
\label{rmk:periodicity}
Since the complex exponential function $k\to e^{-2\pi i\frac{j}{N}k}$ is a $N-$periodic function for any integer $j$, we set 
\[
\hat{\mu}_{k}:=\hat{\mu}_{mod_N(k)}
\]
for any $k \in \mathbb{Z}$, where $mod_N(k)$ is the $N-$modulo operation. In particular, $\hat{\mu}_{-k}=\hat{\mu}_{N-k}$ for any $k\in\{0,\dots,N-1\}$.
\end{remark}

\begin{remark}
\label{rmk:DFTlinearapp}
The DFT of a discrete measure can be expressed as a linear map:
\begin{equation}
     (\hat{\mu}_{0},\dots,\hat{\mu}_{N-1})=\Omega \cdot (\mu_{0},\dots,\mu_{N-1}),
\end{equation}
where $\Omega$ is the $N\times N$ matrix defined as
\begin{equation}
\label{eq:omega_matrix}
    \Omega:=\begin{bmatrix}
\omega_{0,0} & \omega_{0,1} & \dots & \omega_{0,N-1}\\
\omega_{1,0} & \omega_{1,1} & \dots & \omega_{1,N-1}\\
\dots        & \dots        & \dots & \dots       \\
\omega_{N-1,0} & \omega_{N-1,1} &\dots  & \omega_{N-1,N-1}
\end{bmatrix},
\end{equation}
and $\omega_{k,j}:=e^{-2\pi i \frac{j}{N}k}$. Since the matrix $\Omega$ is invertible, the DFT is a bijective function. 
\end{remark}

% \begin{proposition}
% \label{prop:conjfreq}
% Let $\mu$ be a discrete measure; then
% \begin{equation}
%     \label{eq:coniug}
%     \hat{\mu}_{N-k}=\overline{\hat{\mu}_{k}}, \quad \quad k \in \{1,\dots,N-1\},
% \end{equation}
% where $\overline{\hat{\mu}_{k}}$ denotes the complex conjugate of $\hat{\mu}_{k}$. In particular, we have
% \begin{equation}
%      \big|\hat{\mu}_{N-k}\big|^2=\big|\hat{\mu}_{k}\big|^2, \quad \quad k \in \{1,\dots,N-1\}.
% \end{equation}
% \end{proposition}

\noindent We now introduce the $1,2-$Periodic Fourier-based Metric \cite{auricchio2020equivalence}.

\begin{definition}%[$1,2-$Periodic Fourier-based Metric, \cite{auricchio2020equivalence}]
\label{def:periodic_fourier}
Let $\mu$ and $\nu$ be two discrete measures over $I_N$. The $1,2-$Periodic Fourier-based Metric is defined as
\begin{equation}
    \label{eq:1,2PFBM}
    f_{1,2}^2\mn:=\int_{[0,1]}\dfrac{\big|\sum_{j=0}^{N-1}(\mu_j-\nu_j)e^{-2\pi i j k}\big|^2}{|k|^2}dk.
    %f_{1,2}^2\mn:=\int_{[0,1]}\dfrac{\big|\hat{\mu}_k - \hat{\nu}_k\big|^2}{|k|^2}dk.
\end{equation}
\end{definition}
In \cite{auricchio2020equivalence}, it is proved that the integral in $\eqref{eq:1,2PFBM}$ converges for any pair of probability measures $\mu$ and $\nu$, and that $f_{1,2}$ is equivalent to $W_1$.

\section{The Fourier Discrepancy Function}
\label{thisd}

In this section, we introduce the Fourier Discrepancy function, inspired by 
\eqref{eq:1,2PFBM}.  

We compare 
the Fourier Discrepancy Function with other discrepancies, and we show with an example its ability to take into account the geometry of the underlying space.
Then, we prove that the Fourier Discrepancy Function is convex, and we provide the explicit formula for the gradient and the Hessian matrix of its corresponding loss function. 
Finally, we present a statistical model with Gaussian noise in the space of frequencies, in which the minimisation of the Fourier Discrepancy is equivalent to the maximisation of the likelihood.
% Finally, we show how the minimisation of the Fourier Discrepancy is equivalent to the maximisation of the likelihood 
% when we deal with Gaussian noise in the space of frequencies.

\begin{remark}
% In what follows,
Herein, we only consider one-dimensional discrete measures, but all the results may be extended to a multi-dimensional setting.

Moreover, for the sake of simplicity, we assume that $N$ is an even number.
%\code{La frase che segue la toglierei}
%Notice that, in many applications, the grid cardinality is usually a power of $2$ \genna{[cit]}.
% \genna{However, all the results hold for any multi-dimensional discrete measure.}
%we have only considered discrete measures over a subset of $\mathbb{R}$. When we move to higher dimensions, we can still recover the regularity results through similar arguments. 
% In particular, since the Fourier Discrepancy function can be express as a bi-linear form, its gradient and its hessian are related to the tensor inducing this bi-linear form.
\end{remark}

Since $\hat{\mu}_k = \overline{\hat{\mu}_{N-k}},$  
we have 
\begin{equation}
\label{rmk:conjiug_coef}
    \big|\hat{\mu}_k-\hat{\nu}_k\big|=\big|\hat{\mu}_{N-k}-\hat{\nu}_{N-k}\big|,
\end{equation}
which means that the $k$-th discrete frequencies give us the same information of the $(N-k)$-th ones. 
%\zambo{For this reason, for any $k \in \{1,\dots,\frac{N}{2}-1\}$, we only consider  the $k$-th frequency, and we take half of the $\frac{N}{2}$-th frequency.}
Therefore, we only consider the first $\frac{N}{2}-1$ frequencies and we take half of the $\frac{N}{2}$-th frequency. We then propose the following discrete version of the metric in \eqref{eq:1,2PFBM}.

% \zambo{Sì io la toglierei, il motivo tra l'altro non è che facciamo la media con la frequenza 0 (almeno per come la interpreto io): è che per ogni coppia di frequenze prendiamo uno solo degli elementi della coppia; per quella $\frac{N}{2}$, essendo da sola, ha senso prenderne metà}
% where $\lfloor a \rfloor$ denotes the greatest $m\in \mathbb{Z}$ such that $m\leq a$. 
% Moreover,
% Finally, we observe that, once we fix $N$, the term $\sqrt{N}$ in equation \eqref{eq:metrica_discretizzata} is a constant, and thus we omit it from the definition of our discrepancy function.
% \end{remark}

%\code{Nella definizione qui sotto bisogna oltre che a normalizzare forse pensare di mettere un 2 per prendere tutte le frequnze che forse poi ci serve nei tight bounds}
\begin{definition}%[Fourier Discrepancy function]
\label{def:FLF}
We define the Fourier Discrepancy function \\$\ff:\PP(I_N)\times\PP(I_N)\to [0,+\infty)$ as 
% \begin{equation}
% \label{eq:discreteF}
%  \ff\mn^2:=\sum_{k=1}^{\smallceilN-1}\dfrac{|\hat{\mu}_k-\hat{\nu}_k|^2}{|k|^2}+\dfrac{|\hat{\mu}_{\frac{N}{2}}-\hat{\nu}_{\frac{N}{2}}|^2}{|N|^2}.
% \end{equation}

\begin{align}
\nonumber\ff^2\mn &:=
\sum_{k=1}^{\smallceilN-1}\dfrac{|\hat{\mu}_k-\hat{\nu}_k|^2}{|k|^2}+
\dfrac{\big|\frac{1}{2}\big(\hat{\mu}_{\frac{N}{2}}-\hat{\nu}_{\frac{N}{2}}\big)\big|^2}{|\frac{N}{2}|^2} \\
\label{eq:discreteF}&= \sum_{k=1}^{\smallceilN-1}\dfrac{|\hat{\mu}_k-\hat{\nu}_k|^2}{|k|^2}+\dfrac{|\hat{\mu}_{\frac{N}{2}}-\hat{\nu}_{\frac{N}{2}}|^2}{|N|^2}.
\end{align}

\end{definition}
% \noindent Notice that the term $\frac{1}{|k|^2}$ in \eqref{eq:discreteF} penalises the differences between higher frequencies.\\

%\noindent Through arguments similar to the ones exposed in \cite{auricchio2020equivalence}, it is possible to prove the following result.
% it is possible to see that this discretization is still a distance. Moreover, it also inherits the equivalence with the $W_1$ distance. 
\begin{remark}
\label{remark_dist}
The function $\ff$ is a distance on $\PP(I_N)$. 
Moreover, the following holds:
\begin{equation}
\label{eq:W_1equiv}
    \frac{1}{N} C_1 \cdot W_1 \le \mathbb{F} \le C_2 \cdot W_1,
\end{equation}
where $C_1, C_2$ are positive constants that do not depend from $N$. 
This follows from the equivalence between the Fourier-based metric and the Wasserstein distance \cite{auricchio2020equivalence}.
\end{remark}

% \label{remark_equiv}
% %From the equivalence between the Fourier-based metric and the Wasserstein distance \cite{auricchio2020equivalence}, follows that
 
% \end{remark}

% \begin{lemma}
% \label{lemma_eq}
% The function $\ff$ defined in \eqref{eq:discreteF} is a distance on $\PP(I_N)$. Moreover, it is equivalent to the Wasserstein distance $W_1$, and
% \begin{equation}
% \label{eq:W_1equiv}
%     \frac{1}{N} C_1 \cdot W_1 \le \mathbb{F} \le C_2 \cdot W_1,
% \end{equation}
% where $C_1, C_2$ are positive constants that do not depend from the discretization.
% \end{lemma}

% \zambo{Non so se scriverei che è equivalente alla W1 dato che siamo nel discreto. E poi metterei la proof anche di questa nell'appendice}
% \code{forse per dargli un senso potremmo dire che è equivalente e sta sotto la W1 con una costante che non dipende dalla discretizzazione e che quindi rileva la convergenza, come vediamo nel successivo esempio}
%We highlight that the constant $C_2$ in \eqref{lemma_eq} does not depend on $N$.

\begin{figure}[t!]
    \centering
    \includegraphics[width=0.47\textwidth]{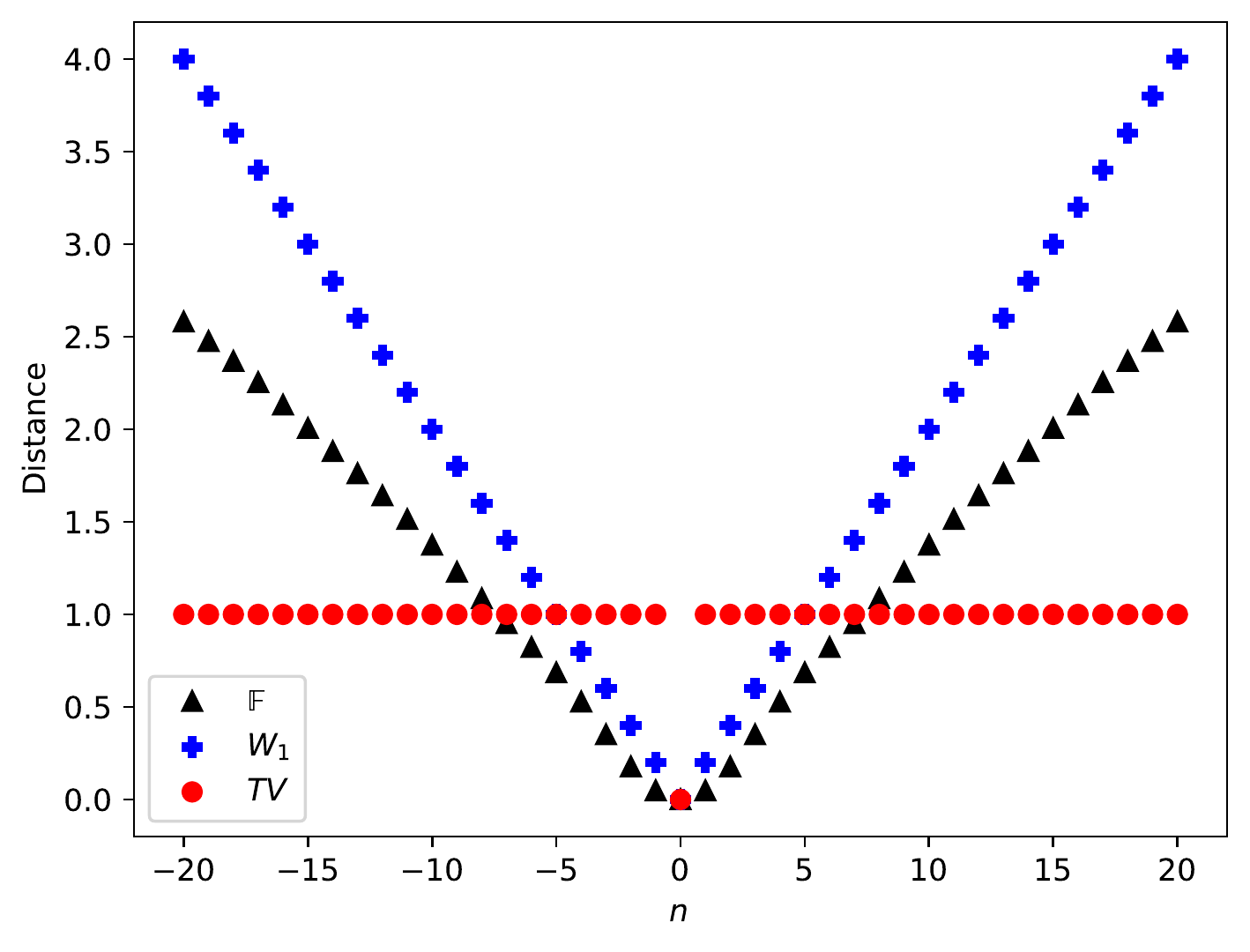}
    \caption{Distance between $\delta_0$ and
    $\delta_n$. We have scaled the distances for visual convenience.}
    \label{fig:delta3}
\end{figure}

\begin{example}
Figure \ref{fig:delta3} shows the behaviours of different discrepancy functions when comparing Dirac's delta distributions. We have omitted the $KL$, since it is always equal to $+\infty$ whenever the supports of the two distributions are disjoint.
We highlight how the Fourier discrepancy, similarly to the $W_1$, is able to take into account the geometry of the underlying space.

\end{example}

To conclude, we provide an upper bound for the Fourier Discrepancy with respect to the Total Variation and the Kullback-Leibler.

\begin{proposition}
\label{prop:tv_ff}
For any pair of probability measures $\mu$ and $\nu$, we have that
\begin{equation}\label{ineqTV}
\ff\mn\leq \frac{2}{\sqrt{6}} \pi \, TV\mn.
\end{equation}
\end{proposition}

The proof is reported in Appendix \ref{app:tv}. 

\begin{remark}
Using the Pinsker's inequality \cite{kullback1967lower}, we also obtain the following bound:
\begin{equation*}
\ff\mn \leq \frac{\pi}{\sqrt{3}} \sqrt{KL(\mu||\nu)}.
\end{equation*}
\end{remark}

\subsection{Analytical properties}

In what follows, we study the analytical properties of the Fourier Discrepancy Function.\\
% \zambo{TOGLIEREI LA PROSSIMA FRASE, PERCHE' LO ABBIAMO GIA' DETTO ALL'INIZIO DELLA SEZIONE:}

% We show that it is convex and we provide the explicit formulae for the gradient and Hessian matrix of its corresponding loss function.\\
%Remarkably, these properties are preserved if the Fourier Discrepancy is used as a loss function. In this framework, we are able to prove that the function is either convex and twice differentiable.\\

% Let us introduce the following matrix
% \begin{equation}
%     \KK := \diag(b)
% \end{equation}
Let us introduce the matrix $\KK := \diag(b)$,
where the vector $b$ is defined as
% \begin{align}
% \label{eq:vector_b}
% \nonumber    b:=\dfrac{1}{2}\bigg(1,|1|^{-2},&\dots,\bigg|\frac{N}{2}-1\bigg|^{-2},2\big|N\big|^{-2},\\
%     &\bigg|\frac{N}{2}-1\bigg|^{-2},\dots,|1|^{-2}\bigg).
% \end{align}
% \zambo{Stavo provando a vedeer se ci stava in una riga sola.. secondo me a togliere  i -2 agli esponenti dell'ultima e della prima componente ci sta. mi sembra ragionevole così, però forse dobbiamo aggiungere il numero per la ref? tho, guarda che bella.
% }
\small
\begin{equation}
\label{eq:vector_b}
    \displaystyle{b:=\dfrac{1}{2}\bigg(1,1^{-2},\dots,\bigg(\frac{N}{2}-1\bigg)^{-2},\frac{2}{N^2},\bigg(\frac{N}{2}-1\bigg)^{-2},\dots,1^{-2}\bigg)}
\end{equation}
\normalsize
%Thanks to Remark \ref{rmk:conjiug_coef}, we are able to 
We can express the Fourier Discrepancy function as a quadratic form:
\begin{align}
\ff^2(\mu,\nu)
&=(\hat{\mu}-\hat{\nu})^T \KK (\hat{\mu}-\hat{\nu}) \nonumber\\
&=(\mu-\nu)^T\Omega^T \KK \Omega (\mu-\nu),
\end{align}
where $\Omega$ is the DFT matrix defined in \eqref{eq:omega_matrix}. 

%Through this expression of the Fourier Discrepancy, we are able to easily infer the analytical properties of the function by simply studying the matrix $\mathbb{H}:=\Omega^T \KK \Omega$.

We now study the matrix $\mathbb{H}:=\Omega^T \KK \Omega$ to derive the analytical properties of the Fourier Discrepancy.

\begin{proposition}
The matrix $\mathbb{H}$ is positive definite and its eigenvalues
%$\lambda_0,\dots,\lambda_{N-1}$ 
are given by 
\begin{equation*}
    \lambda_i = N \cdot b_i, \qquad i = 0,\dots,N-1,
\end{equation*}
where $b$ is the vector in \eqref{eq:vector_b}.
\end{proposition}
%In particular, the function $\ff^2$ is convex with respect to the measure $\mu-\nu$. 

Since $\mathbb{H}$ is positive definite, there exists a matrix $\mathbb{L}$ such that $\mathbb{L}^T \mathbb{L} = \mathbb{H}$. Therefore, we can write
\[
\ff(\mu-\nu) = \|\mathbb{L}(\mu-\nu)\|.
\]
Since $\ff$ is given by the composition of a linear function with the norm operator, we have the following.

\begin{proposition}
The Fourier Discrepancy is convex in $\mu-\nu$.
\end{proposition}

In many applications, discrepancies are used to evaluate how different a given probability measure is from a target one. An established tool to perform this comparison is the loss function.

For any given $\nu\in\PP(I_N)$, we define the Fourier Loss Function $L_\nu:\PP(I_N)\to [0,\infty)$ as 
\begin{equation}
    L_\nu(\mu):=\ff^2(\mu,\nu).
\end{equation}
We are able to explicitly express the gradient and the Hessian matrix of this function.

\begin{proposition}
\label{prop:gradfour}
For any probability measure $\nu$, the function $L_\nu$ is twice differentiable. Moreover, its gradient and Hessian matrix are expressed through the explicit formulae:
\begin{equation}
\label{eq:fouriergradient}
  (\nabla L_\nu)_l(\mu)= \frac{\partial L_\nu}{\partial \mu_l}(\mu)=  2\sum_{j=0}^{N-1} (\mu_j- \nu_j) \cdot \Real \left( \hat{b}_{j-l}\right)
\end{equation}
and
\begin{equation}
\label{eq:fourierhessian}
    (HL_\nu)_{h,l}(\mu)=\frac{\partial^2 L_\nu}{\partial \mu_h \partial \mu_l}(\mu) = 2\Real \left( \hat{b}_{h-l}\right),
\end{equation}
where $\hat{b}$ is the Fourier Transform of the vector $b$. 

In particular, $L_\nu$ is a convex function for any $\nu\in\PP(I_N)$.
\end{proposition}

\subsection{Statistical interpretation}

We now show how the minimisation of the Fourier Discrepancy is related to the maximum likelihood estimator in classification models with a random noise. This is a classic framework in machine learning, where we often assume the existence of an underlying probabilistic model that generates the data \cite{bengio2017deep,murphy2012machine}. This model is typically expressed as
\begin{equation}\label{model prob interp}
  y_i=f(x_i;\theta)+\epsilon_i, \quad\quad i=1,\dots,m,  
\end{equation}
where $(x_1,y_1),\dots,(x_m,y_m)$ are the data, $\epsilon_1,\dots,\epsilon_m$ are i.i.d. random noises, $f$ is a function that specifies the model structure, and $\theta$ is the parameter that has to be optimised.

Let us suppose that, for every $i=1,\dots,m$, $\hat{\epsilon_i} \sim \mathcal{CN}(0, \Sigma)$, where  $\mathcal{CN}(0,\Sigma)$ is the circularly-symmetric complex normal distribution with zero mean and covariance matrix $\Sigma$, defined as $\Sigma := \diag\big( 2 \sigma^2 \beta\big)$, for some $\sigma > 0$, and where $\beta$ is given by
\[
\beta:=\bigg(0,1^2,\dots,\bigg|\frac{N}{2}-1\bigg|^2,\bigg|\frac{N^2}{2}\bigg|,\bigg|\frac{N}{2}-1\bigg|^2,\dots,1^2\bigg).
\]
% \diag\big(0, 1\sigma^2, 4\sigma^2, \dots, (\frac{N}{2}-1)^2 \sigma^2,N^2\sigma^2\big)$,

For a complete discussion on complex normal distributions, we refer to \cite{goodman1963}.

%\genna{riguardare} We recall the basic notions about complex normal distributions in the additional material; for a complete discussion, we refer to \cite{goodman1963}. 

The likelihood of the observations $(x,y) = (x_i, y_i)_{i=1,\dots,m}$ is then given by
\begin{align}\label{likelihood noise on frequencies}
\nonumber \mathbb{P}(y|x, \theta) 
&= \prod_{i=1}^m \mathbb{P}(y_i = f_i + \epsilon_i) \\
&= \prod_{i=1}^m \mathbb{P}(\hat{y_i} = \hat{f_i} + \hat{\epsilon_i}) \nonumber \\ \nonumber
&= \const \cdot \prod_{i=1}^m\exp \left[ - \frac{1}{2} (\hat{y_i} - \hat{f_i})^H \Sigma^{-1} (\hat{y_i} - \hat{f_i}) \right]  \\ 
&= \const \cdot \exp \Bigg[ - \frac{1}{2\sigma^2} \sum_{i=1}^m
\ff^2(y_i,f_i)
%\bigg( \sum_{k=1}^{\smallceilN-1} \frac{|\hat{y_i}(k) - \hat{f_i}(k)|^2}{k^2}\\
%&\quad\quad\quad\quad\quad\;+ \frac{|\hat{y_i}(\frac{N}{2}) - \hat{f_i}(\frac{N}{2})|^2}{N^2} \bigg)
\Bigg],
% \cdot \mathbbm{1}_{\{ \hat{y_i}(0) = \hat{f_i}(0) \, \forall i \}},
\end{align}
\noindent provided that the mass of $y_i$ is the same as $f_i$, for every $i = 1,\dots,m$. The vector $(\hat{y_i} - \hat{f_i})^H$ denotes the conjugate transpose of $(\hat{y_i} - \hat{f_i})$ and $C$ is a positive constant that does not depend on the data.

By taking the logarithm in \eqref{likelihood noise on frequencies}, we obtain the following result.

\begin{theorem}
Let us consider a model of the form \eqref{model prob interp}, where $(x_1,y_1),\dots,(x_m,y_m)$ are the data and $\epsilon_1,\dots,\epsilon_m$ are distributed as described above. Then, the value of $\theta$ maximising the likelihood of the data is the one minimising the Fourier Discrepancy
\begin{equation*}
    \frac{1}{m} \sum_{i=1}^m L_{\delta_{y_i}}\left( f(x_i;\theta) \right).
\end{equation*}
\end{theorem} 

Notice that the structure of the covariance matrix $\Sigma$ measures how the error on the $k^{th}$ frequency is weighted.
As the variance grows, we are more willing to accept discrepancies between the real value of the frequency and the predicted one. In particular, for $k=0$, we have a null variance Gaussian (i.e. a Dirac's delta). Therefore the model does not admit any error on the null frequency: $\mu$ and $\nu$ must have the same mass.

% Notice that the structure of the covariance matrix $\Sigma$ measures how the error on the $k^{th}$ frequency is weighted.
% As the variance grows we are more willing to accept discrepancies between the real value of the frequency and the predicted one. In particular, for $k=0$, we have a null variance Gaussian (i.e. a Dirac's delta), therefore the model does not admit any error on the null frequency: $\mu$ and $\nu$ must have the same mass.

%%%%%%%%%%%%%%%%%%%%%%%%%%%%%%%%%%%%%%%%%%%%%%%%%%%%%%%%%%%%%%%%%%%%%%%%%%%%%%%%%%%%%%%%%%%%%%%%%%%%%%%%%%%%%%%%%%

\section{Tight Bounds} \label{section tight bounds}

In this section, 
we study the tight bounds for the Fourier Discrepancy in terms of the Total Variation distance. 

A first result is given in Proposition \ref{prop:tv_ff}.
However, what we aim to find are the lower and upper tight bounds, respectively $C_L(\theta)$ and $C_U(\theta)$, defined, for any given $\theta \in (0,1]$, as
\begin{align}
C_L(\theta) := \inf_{\mu, \nu: TV(\mu,\nu)=\theta}   \mathbb{F}(\mu,\nu), \label{eq:lwr_bnd} \\
C_U(\theta) := \sup_{\mu, \nu: TV(\mu,\nu)=\theta}   \mathbb{F}(\mu,\nu). \label{eq:upp_bnd}
\end{align}

Due to the linearity of the DFT, we have that
\begin{equation*}
\ff\mn^2=\sum_{k=1}^{\smallceilN}\dfrac{|\widehat{(\mu - \nu)}_k|^2}{|k|^2},
\end{equation*}
% and moreover $TV(\mu,\nu)=\frac{1}{2}\sum_{i=0}^{N-1}|\mu_i-\nu_i|$,
we then set
$\Delta:=\mu-\nu$ and express both $TV$ and $\mathbb{F}$ as functions of $\Delta$, rather than $\mu$ and $\nu$.

%In particular, we can rewrite \eqref{eq:max_prob} as
% \[
% C_{UB}=\sup_{\substack{\Delta=\mu-\nu: \\ 
% %\text{where}\;
% \mu,\nu\in\PP(X), \,\mu \neq \nu } } \dfrac{\mathbb{F}(\Delta)}{||\Delta||_{L^{1}}},
% \]

We now introduce and study the space of null sum measures.

\begin{definition}
We say that a real measure $\Delta$ is a null sum measure if
\[
\sum_{i=0}^{N-1}\Delta_i=0.
\]
We denote by $\Theta$ the set of all the null sum measures.
\end{definition}

Given any pair of probability measures $\mu$ and $\nu$, their difference is 
% obviously 
a null sum measure. As the following result shows, up to a multiplicative constant, the converse is also true.

\begin{proposition}\label{proposition delta 1}
Given any non-zero $\Delta\in\Theta$ and $\theta \in (0,1]$, there exists $C > 0$ and a pair of probability measures $(\mu,\nu)$ such that 
\[
\mu-\nu = C \cdot \Delta \quad \text{and} \quad TV(\mu,\nu)=\theta.
\]
\end{proposition}
 
 \begin{proof}
 
 Let $C := \frac{\theta}{TV(\Delta)}$ and $\widetilde{\Delta} := C \cdot \Delta$, which are well-defined since $TV(\Delta) \neq 0$ for any non-zero $\Delta$.
 
 Then, for the $1-$homogeneity of $TV$, we have that $TV(\widetilde{\Delta}) = \frac{\theta}{TV(\Delta)}\cdot{TV(\Delta)} = \theta$.\\
 
Let $\widetilde{\mu}$ and $\widetilde{\nu}$ be, respectively, the positive and negative part of $\widetilde{\Delta}$.
Therefore, $\widetilde{\Delta} = \widetilde{\mu} - \widetilde{\nu}$ and $\widetilde{\mu}_i, \widetilde{\nu}_i \ge 0$ for any $i$.

We have that
 \begin{equation}\label{proof delta 1}
 2 \theta = \sum_i |\widetilde{\Delta}_i| = \sum_i \widetilde{\mu}_i + \sum_i \widetilde{\nu}_i,
 \end{equation}
and moreover, since $\widetilde{\Delta}$ is a null sum measure:
 \begin{equation} \label{proof delta 2}
 0 = \sum_i \widetilde{\Delta}_i = \sum_i \widetilde{\mu}_i - \sum_i \widetilde{\nu}_i. 
 \end{equation}

From \eqref{proof delta 1} and \eqref{proof delta 2} follows easily that $\sum_i \widetilde{\mu}_i = \sum_i \widetilde{\nu}_i = \theta$.
 
We now define 
\[\mu := \widetilde{\mu} + (1-\theta)\delta_0, \quad \nu := \widetilde{\nu} + (1-\theta)\delta_0.\]

We have that $\mu$ is a probability measure since $\mu_i \ge 0$ for any $i$ and $\sum_i \mu_i = \sum_i \widetilde{\mu}_i + (1-\theta) = 1$. The same holds for $\nu$.

Moreover, $\mu - \nu = \widetilde{\Delta}$, hence
$TV(\mu,\nu) = TV(\widetilde{\Delta}) = \theta$.
 \end{proof}
 
\begin{remark}\label{remark delta}
Thanks to Proposition \ref{proposition delta 1}, and for the $1$-homogeneity of $\ff$, we have that
\begin{align}\label{inf rem delta}
C_L(\theta) &= \inf_
{\substack{\Delta \in \Theta: \\  \Delta \neq 0} }
%{\Delta \in \Theta: \, \Delta \neq 0}
\; \ff\left( \frac{\theta}{TV(\Delta)} \,\Delta \right) \nonumber \\
&= \theta \cdot \inf_
{\substack{\Delta \in \Theta: \\  \Delta \neq 0} }
%{\Delta \in \Theta: \, \Delta \neq 0}
\,\frac{\ff(\Delta)}{TV(\Delta)},
\end{align}
and, analogously,
\begin{equation}\label{sup rem delta}
C_U(\theta) = \theta \cdot \sup_
{\substack{\Delta \in \Theta: \\  \Delta \neq 0} }
%{\Delta \in \Theta: \, \Delta \neq 0} 
\,\frac{\ff(\Delta)}{TV(\Delta)}. 
\end{equation}
\end{remark}

\subsection{Lower tight bound}

%Let $\Delta$ be a non-zero null sum measure,
%Let us take a non-zero null sum measure $\Delta$,
%Let us denote by $\omega_k$ the $(k+1)$-th row of the matrix $\Omega$, that is
Let us define the complex vector $\omega_k \in \mathbb{C}^N$ as  
\[
\omega_{k}=\Big(e^{i\frac{2\pi k}{N}0},e^{i\frac{2\pi k}{N}1},\dots,e^{i\frac{2\pi k}{N}(N-1)} \Big).
\]
Since $\{\omega_k\}_{k=0,\dots,N-1}$ is an orthogonal basis of $\mathbb{C}^n$ \cite{britanak2001transform}, for any $\Delta \in \Theta$ there exists a unique $N$-tuple of complex coefficients $\big(\lambda^{(k)}\big)_k$ such that
\[
\Delta=\sum_{k=0}^{N-1}\lambda^{(k)}\omega_k.
\]
%Since $\Delta$ is a null sum measure, we infer $\lambda^{(0)}=0$ 
%\zambo{Notice that $\lambda^{(0)} = 0$ because... (forse si può non mettere, o mettere dopo)}. 
We define 
\begin{equation}\label{def XI}
\Xi := \Big\{\Delta \in \Theta: \; \sum_{k=0}^{N-1} |\lambda^{(k)}| = 1\Big\}.
\end{equation}
From \eqref{inf rem delta}, and for the $1$-homogeneity of both $TV$ and $\mathbb{F}$, we have that:

\begin{align} \label{inf rem delta 2}
C_L(\theta) 
&= \theta \cdot \inf_
{\substack{\Delta \in \Theta: \\  \Delta \neq 0} }
%{\Delta \in \Theta: \, \Delta \neq 0} 
\,\frac{\ff\left(\frac{\Delta}{\sum|\lambda^{(k)}|}\right)}{TV\left(\frac{\Delta}{\sum|\lambda^{(k)}|}\right)} \frac{\sum|\lambda^{(k)}|}{\sum|\lambda^{(k)}|}, \nonumber\\
&= \theta \cdot 
\inf_
{\Delta \in \Xi}
%{\substack{\Delta \in \Theta: \\  \sum |\lambda^{(k)}| = 1} }
%{\Delta \in \Theta: \, \sum |\lambda^{(k)}| = 1}
\,\frac{\ff(\Delta)}{TV(\Delta)}.
\end{align}

% Therefore, we consider only the subset of vector $V_{\Omega}:=\{\omega_{k}\}_{k=2,\dots,N}$. 
% Through this set, we can describe every element of $\Theta$ as a linear combination, that is
% \begin{equation}
% \label{eq:ration}
% \Delta=\sum_{k=2}^{N}\lambda^{(k)}\omega_k
% \end{equation}
% however, due the $0-$homogeneity of the ratio, we will always assume that the coefficients $\lambda^{(k)}$ are such that
% \begin{equation}
%     \label{eq:normalise}
%     \sum_{k=2}^N|\lambda^{(k)}|=1.
% \end{equation}

\begin{lemma}\label{lemma sup TV}
We have that
\begin{equation}
\sup_
{\Delta \in \Xi}
%{\substack{\Delta \in \Theta: \\  \sum |\lambda^{(k)}| = 1} }
\;TV(\Delta) = \frac{N}{2},
\end{equation}
and the supremum is attained at $\Delta = 
\omega_{\frac{N}{2}}$.

% We have that
% \begin{equation}
% TV(\omega_k) = N, \qquad \text{for} k = 0,\dots,N-1,
% \end{equation}
% and for any $\Delta \in \Theta$ with $\sum_{k=0}^{N-1} |\lambda^{(k)}| = 1$:
% \begin{equation}
% TV(\Delta) \le N.
% \end{equation}
% Let $\Delta$ be a null sum measure such that $\sum_{k=0}^{N-1} |\lambda^{(k)}| = 1$. Then
% \begin{equation}
% TV(\Delta) \le TV(\omega_k)
% \end{equation}
\end{lemma}

\begin{proof}
Since $|(\omega_k)_j| = |e^{i\frac{2\pi jk}{N}}| = 1$ for all $j$ and $k$, we have that $TV(\omega_k) = \frac{N}{2}$. Then, for any $\Delta \in \Theta$ such that $\sum |\lambda^{(k)}| = 1$, we have
\small
\begin{align*}
TV(\Delta)&=TV\Big(\sum_{k=0}^{N-1} \lambda^{(k)}\omega_k\Big)
=\frac{1}{2}\sum_{j=0}^{N-1}\Big| \sum_{k=0}^{N-1} \lambda^{(k)}(\omega_k)_j\Big|\\
&\leq \frac{1}{2} \sum_{j=0}^{N-1}\sum_{k=0}^{N-1} \Big| \lambda^{(k)}(\omega_k)_j \Big|
=\frac{1}{2}\sum_{k=0}^{N-1} |\lambda^{(k)}|\sum_{j=0}^{N-1}|(\omega_k)_j |\\
&= \frac{N}{2} \sum_{k=0}^{N-1}|\lambda^{(k)}| = \frac{N}{2}.
\end{align*}
\normalsize
Finally, notice that $\omega_{\frac{N}{2}} \in \Theta$ since $\big(\omega_{\frac{N}{2}}\big)_j = e^{i \pi j} = (-1)^j$, therefore $\omega_{\frac{N}{2}}$ is real and $\sum_{j=0}^{N-1} \big(\omega_{\frac{N}{2}}\big)_j = 0$.
\end{proof}

\begin{lemma}
\label{pr:four_ellipse}
For any $\Delta\in\Theta$, the Fourier Discrepancy is given by
\begin{equation}
    \label{eq:four_decomp}
\ff^2(\Delta)= N^2 \bigg(\sum_{k=1}^{\frac{N}{2}-1}
\dfrac{|\lambda^{(k)}|^2}{k^2}+\dfrac{|\lambda^{(\frac{N}{2})}|^2}{|N|^2}\bigg).
\end{equation}
\end{lemma}

\begin{proof}
For any $j = 0,\dots,N-1$, we have that the DFT of $\omega_j$ is given by
\begin{align*}
\widehat{(\omega_j)}_k = \sum_{l=0}^{N-1} e^{-i\frac{2\pi}{N}lk} (\omega_j)_l 
= \sum_{l=0}^{N-1} e^{-i\frac{2\pi}{N}l (k-j)} = N \delta_{k-j}.
\end{align*}

Hence, for the linearity of the DFT: 
\begin{align*}
\widehat{\Delta}_k = \sum_{j=0}^{N-1} \lambda^{(j)} \widehat{(\omega_j)}_k 
= N \sum_{j=0}^{N-1} \lambda^{(j)} \delta_{k-j}
= N \lambda^{(k)}.
\end{align*}
\end{proof}

\begin{lemma} \label{lemma inf F}
We have that
\begin{equation*}
\inf_{\Delta \in \Xi} \;\ff(\Delta) = 1,
\end{equation*}
and the infimum is attained at $\Delta = \omega_{\frac{N}{2}}$.
\end{lemma}

\begin{proof}

Let $\Delta \in \Xi$. Then $\lambda^{(0)} = \sum_j \Delta_j = 0$. 
Moreover, since $\Delta$ is real, we have that $\widehat{\Delta}_k = \overline{\widehat{\Delta}_{N-k}}$ for any $k=1,\dots,N-1$, hence $|\lambda^{(k)}| = |\lambda^{(N-k)}|$.
%Therefore: \[2 \sum_{k=1}^{\frac{N}{2}-1} |\lambda^{(k)}| + |\lambda^{\frac{N}{2}}| = 1.
%\]
If we define
\[
\gamma_j := 
\begin{cases}
2 |\lambda^{(j)}| \quad\;\;\;\, \text{for}\; j=1,\dots,\frac{N}{2}-1,\\
|\lambda^{(\frac{N}{2})}| \qquad  \text{for} \; j=\frac{N}{2},
\end{cases}
\]
from \eqref{eq:four_decomp} we obtain 
\[\ff^2(\Delta) = \Big(\frac{N}{2}\Big)^2 \sum_{k=1}^{\frac{N}{2}} \frac{\gamma_k^2}{k^2},
\]
while the constraint \eqref{def XI} is written as
\[
\sum_{j=1}^{\frac{N}{2}} \gamma_j = 1.
\]

It is easy to see that the minimum is achieved when $\gamma_{\frac{N}{2}}=1$ and $\gamma_j = 0$ for $j=1,\dots,\frac{N}{2}-1$. Therefore $\Delta = \omega_{\frac{N}{2}} \in \Xi$, and $\ff(\Delta)=1$.
%, which is a feasible solution thanks to Lemma \ref{lemma sup TV}.

% Thanks to formula \eqref{eq:four_decomp}, minimising the Fourier Discrepancy over the set $\Xi$ is equivalent to minimising \eqref{eq:four_decomp}, with the constraint
% \[
% \sum_{k=1}^{N-1}|\lambda^{(k)}|=1,
% \]
% \genna{which, according to relation \ref{rmk:conjiug_coef}, is equivalent to
% \[
% \sum_{k=1}^{\frac{N}{2}-1}2|\lambda^{(k)}|+|\lambda^{(\frac{N}{2})}|=1.
% \]}
% It is easy to see that the minimum is achieved when $|\lambda^{(\frac{N}{2})}|=1$ and $|\lambda^{(k)}|=0$ if $k\neq \frac{N}{2}$, which, from identity \ref{rmk:conjiug_coef}, allows us to infer $\lambda^{(k)}=0$ for any $k\neq \frac{N}{2}$. 

% To conclude notice that if $\lambda^{(\frac{N}{2})}=1$, the measure $\omega_{\frac{N}{2}}$ belongs to $\Xi$ and, hence, is a minimiser.

\end{proof}

Combining \eqref{inf rem delta 2} with Lemma \ref{lemma sup TV} and Lemma \ref{lemma inf F}, we infer that the lower tight bound is attained at $\Delta = [-1,1,-1,1,\dots,1]$.
Thanks to Proposition \ref{proposition delta 1}, we can conclude with the following theorem.

\begin{theorem}
The lower tight buond $C_L(\theta)$ is given by 
\begin{equation}
C_L(\theta) = \frac{2 \theta}{N},
\end{equation}
and is attained at 
\begin{align*}
\mu = \frac{2 \theta}{N} [1,0,1,0,\dots,0] + (1-\theta)\delta_0, \\
\nu = \frac{2 \theta}{N} [0,1,0,1,\dots,1] + (1-\theta)\delta_0.
\end{align*}

\end{theorem}

\subsection{Upper tight bound}

First, we introduce a suitable class of null sum measures.
%that allows us to describe every $\Delta \in \Theta$ through a linear combination. 

\begin{definition}
For any $i, j \in \{0,\dots,N-1\},$ we define the measure $\eta_{i,j}$ as
\[
\eta_{i,j}:=\delta_{i}-\delta_{j}.
\]
\end{definition}

\begin{theorem}
\label{theorem:convcomb}
Let $\Delta$ be a null sum measure on $\{0,\dots,N-1\}$. Then, we can express $\Delta$ as $\Delta = TV(\Delta) \cdot \Delta'$, where $\Delta'$ is a convex combination of 
% \begin{equation}
% \Delta = \frac{TV(\Delta)}{2} \sum_k \lambda_k \eta_{i_k,j_k},
% \end{equation}
$\{\eta_{i_k,j_k}\}_{k}$ such that, for any pair $\eta_{i_k,j_k}$ and $\eta_{i_{k'},j_{k'}}$, we have
%$i_k\neq j_{k'}$ whenever $k\neq k'$.
\begin{equation}
    \label{eq:disjoint}
    i_k\neq j_{k'}
    %\qquad \text{for any} \; k \neq k'.
\end{equation}
for any $k\neq k'$.
\end{theorem}

\begin{proof}
Let $\Delta$ be a null sum measure. Without loss of generality, we can reorder the values of $\Delta$ as follows:
\[
\Delta = (\alpha_1,\dots,\alpha_r,-\beta_1,\dots,-\beta_l,0,\dots,0),
\]
where $r+l\leq N$, $\alpha_i,\,\beta_j> 0$, $\alpha_{i}\leq\alpha_{i+1}$, $\beta_{j}\leq\beta_{j+1}$, for any $i$ and $j$, and
%Notice that 
$\sum \alpha_i = \sum \beta_j$.

Without loss of generality, we  assume that
\[
\alpha_1\leq \beta_1.
\]
Hence, we can write
\[
\Delta=\alpha_1\eta_{0,r}+\Delta^{(1)},
\]
where
\begin{align*}
    \Delta^{(1)}&=(0,\alpha_2^{(1)},\dots,\alpha^{(1)}_r,-\beta^{(1)}_1,\dots,-\beta^{(1)}_l,0,\dots,0)\\
    :&=(0,\alpha_2,\dots,\alpha_r,-(\beta_1-\alpha_1),-\beta_2,\dots,-\beta_l,0,\dots,0).
\end{align*}
Next, we compare $\alpha^{(1)}_2$ and $\beta^{(1)}_1$ and repeat the process until every entry vanishes. At the end, we find
\begin{align} \label{decomposition Delta}
\Delta
&=\lambda_1\eta_{0,r}+\dots+\lambda_k \eta_{r-1,N-1} \nonumber \\
&=: \sum_{k}\lambda_k\eta_{i_k,j_k}.
\end{align}

Notice that each $\eta_{i,j}$ in \eqref{decomposition Delta} is such that $i<r$ and $j\geq r$ by construction, which implies condition \eqref{eq:disjoint}.

Since by hypothesis, for any $l=0,\dots,N-1$, all the $l$-th entries $(\eta_{i_k,j_k})_i$ have the same sign, we can write
\[
|\Delta_l|
=\Big|\sum_{k}\lambda_k(\eta_{i_k,j_k})_l\Big|
=\sum_{k}\lambda_k|(\eta_{i_k,j_k})_l|.
\]

Therefore:
\begin{align*}
    TV(\Delta) 
    &= \frac{1}{2} \sum_{l}|\Delta_l| 
    =\frac{1}{2}\sum_{l}\sum_{k}\lambda_k|(\eta_{i_k,j_k})_l|\\
    &=\frac{1}{2}\sum_{k}\sum_{l}\lambda_k|(\eta_{i_k,j_k})_l| \\
    &=\frac{1}{2}\sum_{k}\lambda_k\sum_{l}|(\eta_{i_k,j_k})_l|
    %&=\sum_{k}\lambda_k||\eta_{i_k,j_k}||_{L^1} \\
    =\sum_k\lambda_k,
\end{align*}
since $\sum_{l}|(\eta_{i,j})_l|=2$ for any $i,j$.
To conclude, it suffices to set
% The proof is concluded by setting %$\widetilde{\lambda}_k := \frac{\lambda_k}{TV(\Delta)}$
%, for all $k=0,\dots,N-1$, 
%and 
\begin{equation*}
\Delta' := \frac{1}{TV(\Delta)} \Delta
= \sum_k \widetilde{\lambda}_k \eta_{i_k,j_k},
\end{equation*}
where 
$\widetilde{\lambda}_k := \frac{\lambda_k}{\sum_l \lambda_l} > 0$, and $\sum_k \widetilde{\lambda}_k = 1$.
\end{proof}

\begin{theorem}\label{final theorem tight bounds}

There exist $i^{\star}, j^{\star} \in \{0,\dots,N-1\}$ such that, for any $\theta \in (0,1]$:
\begin{equation}
\theta \cdot \eta_{i^{\star},j^{\star}} = \argmax_{TV(\Delta) = \theta} \, \mathbb{F}(\Delta).
\end{equation}
\end{theorem}

\begin{proof}
First, let us define
\begin{equation}
(i^{\star},j^{\star}) := 
\argmax_{i,j \in \{0,\dots,N-1\}} \mathbb{F}(\eta_{i,j}),
\end{equation}
which exists since the maximum is taken over a finite set.
For any $\theta \in (0,1]$ and any null sum measure $\Delta$ with $TV(\Delta) = \theta$, thanks to Theorem \ref{theorem:convcomb}, we can write $\Delta = \theta \cdot \sum_k \lambda_k \eta_{i_k,j_k}$.

From the $1$-homogeneity and the convexity of $\mathbb{F}$, we obtain: 
\begin{align}
\mathbb{F}(\Delta) &=
%\mathbb{F}(\theta \Delta') = 
%\theta \mathbb{F}(\Delta') =
\mathbb{F}\left(\theta \cdot \sum_k \lambda_k \eta_{i_k,j_k}\right)  
= \theta \cdot  \mathbb{F}\left(\sum_k \lambda_k \eta_{i_k,j_k}\right) \nonumber \\
&\le \theta \cdot \sum_k \lambda_k
\mathbb{F}\left(\eta_{i_k,j_k}\right) 
\le \theta \cdot \sum_k \lambda_k \mathbb{F}\left(\eta_{i^{\star},j^{\star}}\right) \nonumber \\
&= \theta \cdot \mathbb{F}\left(\eta_{i^{\star},j^{\star}}\right) 
= \mathbb{F}\left(\theta \cdot \eta_{i^{\star},j^{\star}}\right).\nonumber
\end{align}
\end{proof}

% \begin{remark}
% This technique is actually rather general, since it may be applied every time we are interested in finding the maximum of a ratio of the form
% \begin{equation}
% \sup_{\mu \neq \nu}\; \frac{f(\mu-\nu)}{TV(\mu,\nu)},
% \end{equation}
% for some $1$-homogeneous and convex function $f$ \zambo{(... giusto? o è anche più generale?)}.
% The above reasoning can be used to show that the supremum is always attained at some pair of measures $\mu=\delta_{i^{\star}}$ and $\nu=\delta_{j^{\star}}$.
% \end{remark}
As a straightforward consequence, we get the following result.
\begin{corollary}
\label{cor:num_fin}
The upper tight bound $C_U(\theta)$ is given by
\begin{equation}
\label{eq:maximizer}
C_U(\theta) = \theta \cdot \mathbb{F}(\eta_{i^{\star},j^{\star}}).
\end{equation}
\end{corollary}

Corollary \ref{cor:num_fin} allows to search for the upper tight bound over a finite set of points.
By explicit computation of the Fourier Discrepancy (see Appendix \ref{app:max}), we have that
\begin{equation*}
\ff^2(\eta_{j,l})
=\sum_{k=1}^{\frac{N}{2}-1}\;\frac{2-2\cos\Big(\dfrac{2\pi |j-l|}{N}k\Big)}{k^2}+\frac{2-2\cos(\pi |j-l|)}{N^2}.
\end{equation*}

Notice that $\ff(\eta_{j,l})$ depends on $j$ and $l$ only through $d:=|j-l|$. Hence, we can further restrict to measures of the form $\eta_{0,d}$, with $d \in \{1,\dots,N-1\}$. 
By studying the derivatives with respect to $d$, it is possible to show that $d^*=\frac{N}{2}$ is a local minimum for  the function $g:[0,N] \to \mathbb{R}$, defined as:
\begin{equation}
\label{eq:min}
g(d):=   \sum_{k=1}^{\frac{N}{2}-1}\;\;\frac{\cos\Big(\dfrac{2\pi d}{N}k\Big)}{k^2}+\frac{\cos(\pi d)}{N^2}.
\end{equation}

We close our paper with the following open conjecture.

\begin{conjecture}
$d^*=\frac{N}{2}$ is
a global minimum for $g$. 
\end{conjecture}

% Our conjecture is that this minimum is also global.
If our conjecture was true, we would have
\[
C_U(\theta) = \theta \cdot \sqrt{\sum_{k=1}^{\frac{N}{2}-1}\frac{2-2(-1)^k}{k^2}+\frac{2-2(-1)^{\frac{N}{2}}}{N^2}}.
\]

% In appendix \ref{app:max}, we show that, the $\eta_{i^{\star},j^{\star}}$ that maximises equation \eqref{eq:maximizer} is related to the minumum of the following function
% \begin{equation}
% \label{eq:min}
%     \sum_{k=1}^{\frac{N}{2}-1}\;\;\frac{\cos\Big(\dfrac{2\pi j}{N}k\Big)}{k^2}+\frac{\cos(\pi j)}{N^2}.
% \end{equation}

% \genna{By studying the derivatives, it is possible to show that $j=\frac{N}{2}$ is a local minimum for the function in \eqref{eq:min}. Our conjecture is that this minimum is also global.}

% \genna{remark su ricerca su di un numero finito di punti? AGGIUSTARE I CONTI SOTTO SUL MINIMO, FATTO MA RICONTROLLARE}
% % \newpage

% % \input{analytical_prop}

\section*{Acknowledgment}
We are deeply indebted to Giuseppe Toscani for several stimulating discussions and valuable suggestions on the Fourier Based Metrics.
%This paper was written within the activities of the GNFM of INDAM. 
The research was partially supported by the Italian Ministry of Education, University and Research (MIUR): Dipartimenti di Eccellenza Program (2018–2022) - Dept. of Mathematics “F. Casorati”, University of Pavia. The PhD scholarship of Andrea Codegoni is funded by Sea Vision S.r.l..

\ifCLASSOPTIONcaptionsoff
  \newpage
\fi

\bibliographystyle{IEEEtran_g}
\bibliography{IEEEexample_g}

% or


\newpage
\appendices

\section{Proof of Proposition \ref{prop:tv_ff}}
\label{app:tv}
\begin{proof}
Let us fix two probability measures $\mu$ and $\nu$ over $I_N$. By definition, we have
\begin{align}
\label{eq:inequastorpia}
    \nonumber\ff\mn^2&=\sum_{k=1}^{\smallceilN-1}\dfrac{\big|\sum_{j=0}^{N-1}(\mu_j-\nu_j)(e^{-2\pi i \frac{j}{N} k})\big|^2}{|k|^2}\\
    \nonumber&\quad\quad\quad+\frac{\big|\sum_{j=0}^{N-1}(\mu_{j}-\nu_{j})(e^{-i\pi j } )\big|^2}{|N|^2}\\
    &\leq \sum_{k=1}^{\smallceilN}\dfrac{\big|\sum_{j=0}^{N-1}|\mu_j-\nu_j||e^{-2\pi i \frac{j}{N} k}|\big|^2}{|k|^2}\\
    \nonumber&= \sum_{k=1}^{\smallceilN}\dfrac{\big|\sum_{j=0}^{N-1}|\mu_j-\nu_j|\big|^2}{|k|^2}\\
    \nonumber&=4TV\mn^2 \sum_{k=1}^{\smallceilN}\dfrac{1}{k^2}\\
    \nonumber&\leq 4TV\mn^2 \sum_{k=1}^{+\infty}\dfrac{1}{k^2}\\
    \nonumber&=\dfrac{4\pi^2}{6}TV\mn^2,
\end{align}
where inequality \eqref{eq:inequastorpia} follows from the fact that $N^{-2}\leq \big(\frac{N}{2}\big)^{-2}$.
The proof is concluded by taking the square root on both sides. 
\end{proof}

\section{Computing $\ff(\eta_{j,l})$}
\label{app:max}
% \zambo{Questa parte l'ho solo copiata, senza modificare nulla}

Let us consider null sum measures of the form $\eta_{l,j}$. We recall that $\eta_{l,j}:=\delta_l - \delta_j$. 
%Without loss of generality, we assume $l<j$. 
Since 
\[
\widehat{\eta_{l,j}}=\Omega\cdot\eta_{l,j},
\]
we have
\begin{equation}
    \widehat{\eta_{l,j}}=\Theta_l-\Theta_j,
\end{equation}
where $\Theta_k$ is the $k-$th column of the matrix $\Omega$. By the definition of $\Omega$ we have
\begin{equation*}
    \Theta_{l}=\Big(  e^{i\frac{2\pi l}{N}0},e^{i\frac{2\pi l}{N}1},\dots,e^{i\frac{2\pi l}{N}(N-1)}     \Big),
\end{equation*}
% where
%               &=\Bigg(  \cos\bigg(\dfrac{2\pi l}{N}0\bigg)+i\sin\bigg(\dfrac{2\pi l}{N}0\bigg),\cos\bigg(\dfrac{2\pi l}{N}1\bigg)+i\sin\bigg(\dfrac{2\pi l}{N}1\bigg),\dots,\\
%               &\quad\quad\quad\quad\cos\bigg(\dfrac{2\pi l}{N}(N-1)\bigg)+i\sin\bigg(\dfrac{2\pi l}{N}(N-1)\bigg)   \Bigg).
therefore, the value $\mathbb{F}(\eta_{l,j})^2$ is then given by
\begin{equation}
\label{eq:sum}
    \mathbb{F}(\eta_{l,j})^2=\sum_{k=1}^{\frac{N}{2}-1}\dfrac{|(\Theta_l-\Theta_j)_k|^2}{k^2}+\dfrac{|(\Theta_l-\Theta_j)_{\frac{N}{2}}|^2}{|N|^2}.
\end{equation}
Let us now compute explicitly $|(\Theta_l-\Theta_j)_k|^2$ for a given $k$. We have
\begin{align*}
    (\Theta_l-\Theta_j)_k&= \cos\bigg(\dfrac{2\pi l}{N}k\bigg)-\cos\bigg(\dfrac{2\pi j}{N}k\bigg)\\
    &\quad+i\sin\bigg(\dfrac{2\pi l}{N}k\bigg) -i\sin\bigg(\dfrac{2\pi j}{N}k\bigg),
\end{align*}

therefore
\begin{align}
    \nonumber|(\Theta_l-\Theta_j)_k|^2&=\Bigg( \cos\bigg(\dfrac{2\pi l}{N}k\bigg)-\cos\bigg(\dfrac{2\pi j}{N}k\bigg)\Bigg)^2\\
    \nonumber&\quad+\Bigg(\sin\bigg(\dfrac{2\pi l}{N}k\bigg) -\sin\bigg(\dfrac{2\pi j}{N}k\bigg)\Bigg)^2\\
    % \nonumber&=\cos\bigg(\dfrac{2\pi l}{N}k\bigg)^2+\cos\bigg(\dfrac{2\pi j}{N}k\bigg)^2\\
    % \nonumber&\quad-2\cos\bigg(\dfrac{2\pi l}{N}k\bigg)\cos\bigg(\dfrac{2\pi j}{N}k\bigg)\\
    % \nonumber&\quad+\sin\bigg(\dfrac{2\pi l}{N}k\bigg)^2+\sin\bigg(\dfrac{2\pi j}{N}k\bigg)^2\\
    % \nonumber&\quad-2\sin\bigg(\dfrac{2\pi l}{N}k\bigg)\sin\bigg(\dfrac{2\pi j}{N}k\bigg)\\
    \nonumber&=2-2\Bigg(\cos\bigg(\dfrac{2\pi l}{N}k\bigg)\cos\bigg(\dfrac{2\pi j}{N}k\bigg)\\
    \nonumber&\quad+\sin\bigg(\dfrac{2\pi l}{N}k\bigg)\sin\bigg(\dfrac{2\pi j}{N}k\bigg)\Bigg)\\
    \label{eq:prost}&=2-2\cos\bigg(\dfrac{2\pi (j-l)}{N}k\bigg),
\end{align}
where the equality in \eqref{eq:prost} comes from the following trigonometric identity:
\[
\cos(\alpha-\beta)=\cos(\alpha)\cos(\beta)+\sin(\alpha)\sin(\beta).
\]

Therefore 
\small
\begin{equation}
\label{F(eta) appendix}
\ff^2(\eta_{j,l})
=\sum_{k=1}^{\frac{N}{2}-1}\;\frac{2-2\cos\Big(\dfrac{2\pi |j-l|}{N}k\Big)}{k^2}+\frac{2-2\cos(\pi |j-l|)}{N^2}.
\end{equation}
\normalsize

\end{document}